\documentclass[sigconf]{acmart}
\usepackage{subcaption}
\usepackage{makecell}
\usepackage{multirow}

\AtBeginDocument{%
  \providecommand\BibTeX{{%
    \normalfont B\kern-0.5em{\scshape i\kern-0.25em b}\kern-0.8em\TeX}}}

\copyrightyear{2024}
\acmYear{2024}
\setcopyright{rightsretained}
\acmConference[DroNet '24 ]{The 10th Workshop on Micro Aerial Vehicle Networks, Systems, and Applications}{June 3--7, 2024}{Minato-ku, Tokyo, Japan}
\acmBooktitle{The 10th Workshop on Micro Aerial Vehicle Networks, Systems, and Applications (DroNet '24 ), June 3--7, 2024, Minato-ku, Tokyo, Japan}\acmDOI{10.1145/3661810.3663464}
\acmISBN{979-8-4007-0656-1/24/06}

\begin{document}

\title{Analyzing Swimming Performance Using Drone Captured Aerial Videos}

\author{Thu Tran}
\orcid{0009-0003-2854-3525}
\email{ndttran.2019@phdcs.smu.edu.sg}
\affiliation{%
  \institution{Singapore Management University}
  \country{}
}


\author{Kenny Tsu Wei Choo}
\email{kenny_choo@sutd.edu.sg}
\orcid{0000-0003-3845-9143}
\author{Shaohui Foong}
\email{foongshaohui@sutd.edu.sg}
\author{Hitesh Bhardwaj}
\email{hitesh_bhardwaj@mymail.sutd.edu.sg}
\author{Shane Kyi Hla Win}
\email{fhlawin_kyi@sutd.edu.sg}
\author{Wei Jun Ang}
\email{weijun_ang@mymail.sutd.edu.sg}
\affiliation{%
  \institution{Singapore University of Technology and Design}
  \country{}
}

\author{Kenneth Goh}
\email{kennethgoh@smu.edu.sg}
\author{Rajesh Krishna Balan}
\email{rajesh@smu.edu.sg}
\affiliation{%
  \institution{Singapore Management University}
  \country{}
}

\renewcommand{\shortauthors}{Tran, et al.}

\begin{abstract}
Monitoring swimmer performance is crucial for improving training and enhancing athletic techniques. Traditional methods for tracking swimmers, such as above-water and underwater cameras, face limitations due to the need for multiple cameras and obstructions from water splashes. This paper presents a novel approach for tracking swimmers using a moving UAV. The proposed system employs a UAV equipped with a high-resolution camera to capture aerial footage of the swimmers. The footage is then processed using computer vision algorithms to extract the swimmers' positions and movements. This approach offers several advantages, including single camera use and comprehensive coverage. The system's accuracy is evaluated with both training and in competition videos. The results demonstrate the system's ability to accurately track swimmers' movements, limb angles, stroke duration and velocity with the maximum error of 0.3 seconds and 0.35~m/s for stroke duration and velocity, respectively.
\end{abstract}




\maketitle

\section{Introduction}
\label{sec:intro}

\begin{figure}[h]
  \centering
  \includegraphics[width=0.8\linewidth]{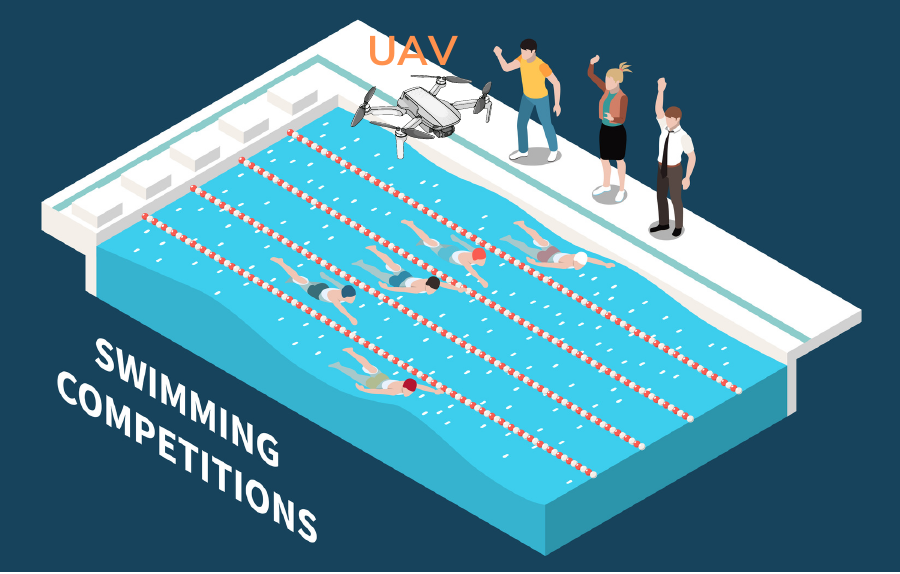}
  \caption{Capturing swimmer footage using a moving UAV}
  \label{fig:uav_intro}
\end{figure}

Video analysis has long been a crucial tool for coaches and athletes in enhancing sports performance. Typically, during training sessions, athletes are recorded, and the resulting footage is later evaluated by coaches and training analysts. However, this process often demands extensive human effort, as annotators are required to manually identify human poses or joint segments in the videos, which can be time-consuming and exhaustive.

In swimming, recent advancements in deep learning (DL), particularly in human pose estimation and video analysis, have enhanced the automated identification of swimming poses. These systems are applicable to both above-water and underwater camera. Above-water cameras are placed on the poolside at a certain height and are mostly used to estimate the speed or the turn of the swimmer as only the swimmer's head is visible in this setting. Underwater cameras can be positioned on the side, in front of the swimmer, or at the pool bottom. Although the first two types of camera setting are widely utilized in swimming analysis due to their high visibility of swimmers' bodies and reduced interference from reflections, and water splashes, they offer only limited coverage, such as focusing solely on the upper body or capturing a single side's limbs. Similarly, while underwater views allow for full observations of swimmers' torso movement, they still require multiple cameras to calculate the athletes' speed.

In this paper, we propose a novel system that leverages a UAV equipped with a high-resolution camera to record aerial footage of swimmers, as shown in Figure \ref{fig:uav_intro}. The captured footage is subsequently analyzed using computer vision algorithms to extract valuable information such as \textbf{swimmers' poses, limb angles, stroke duration, and velocity}. We note that our system is similar to the one proposed in \cite{toyoda2019shooting}, but we do not employ underwater drones. 

The key contributions of our study include:

\begin{itemize}
    \item \textbf{Comprehensive Coverage:} By operating at an optimal height, the UAV guarantees an unobstructed view of the swimmers, providing visibility of all limbs, including both lower and upper ones, as well as capturing movements from both the left and right sides during the strokes.

    \item \textbf{Single Camera Use:} The UAV's movement and aerial perspective provide a sufficient view of the pool, eliminating the need for multiple underwater or poolside cameras to calculate the swimmers' velocity.
\end{itemize}

\section{Related Work}
\label{sec:related}
Regarding above-water cameras, prior studies have utilized specific setups to analyze different aspects of swimming performance. For example, \cite{veiga2013kinematical} positioned four cameras at the public stands, 7~m above and 7~m away from the pool's side. This configuration allowed them to analyze turn performances during backstroke races by estimating the head emersion. Likewise, \cite{gatta2015path} employed a 2-camera system positioned 6~m above and 15~m away from the pool. Their aim was to analyze the head trajectory of elite athletes during a 400-meter front crawl competition, providing insights into the main energy-consuming events that affect their path linearity.

For underwater cameras, \cite{fani2018swim} successfully differentiate the high elbow pose and the dropping pose during the pulling phase of the front crawl stroke with a front-view camera. In addition, \cite{zecha2018kinematic, giulietti2023swimmernet} adapted side-view cameras with novel algorithtms to estimate the pose of athlete swimmers and extract body parts' trajectories with high accuracy.

On the other hand, cameras positioned at the bottom of the pool offer a unique perspective from beneath the water surface, providing less distortion caused by water surfaces and minimal limb dislocation due to lighting refraction. \cite{ukai2013swimoid} developed a buddy robot that can swim under the swimmer, capture footage through its front and rear cameras, and provide feedback through a display mounted on its body. In a more recent study, \cite{cao2023pose} introduced a dataset of swimmers and a novel DL approach to detect joint locations using data collected from underwater cameras. Their dataset demonstrates that, in comparison to a side view angle, a complete view of the swimmer's body can be observed.

\section{System Overview}
\label{sec:system}
\begin{figure*}[h]
  \centering
  \includegraphics[width=0.7\linewidth]{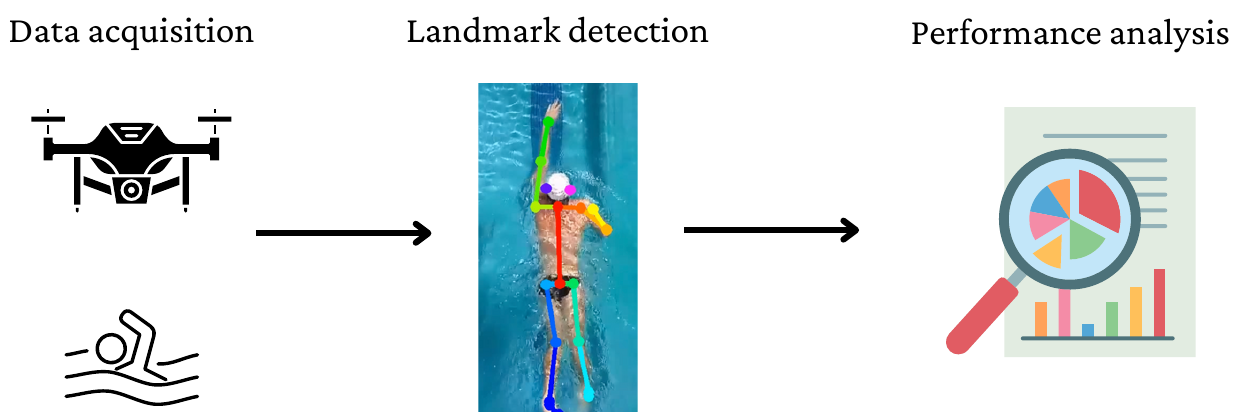}
  \caption{System overview}
  \label{fig:system}
\end{figure*}

Our system has three main components, as illustrated in Figure \ref{fig:system}.

\subsection{Data Acquisition}
We recorded the swimming performances of national athletes, including both their training sessions and competitive events, conducted in a standard Olympic-sized swimming pool.

\begin{figure*}[ht]
\begin{subfigure}[b]{0.4\textwidth}
  \centering
  \includegraphics[width=1.0\linewidth]{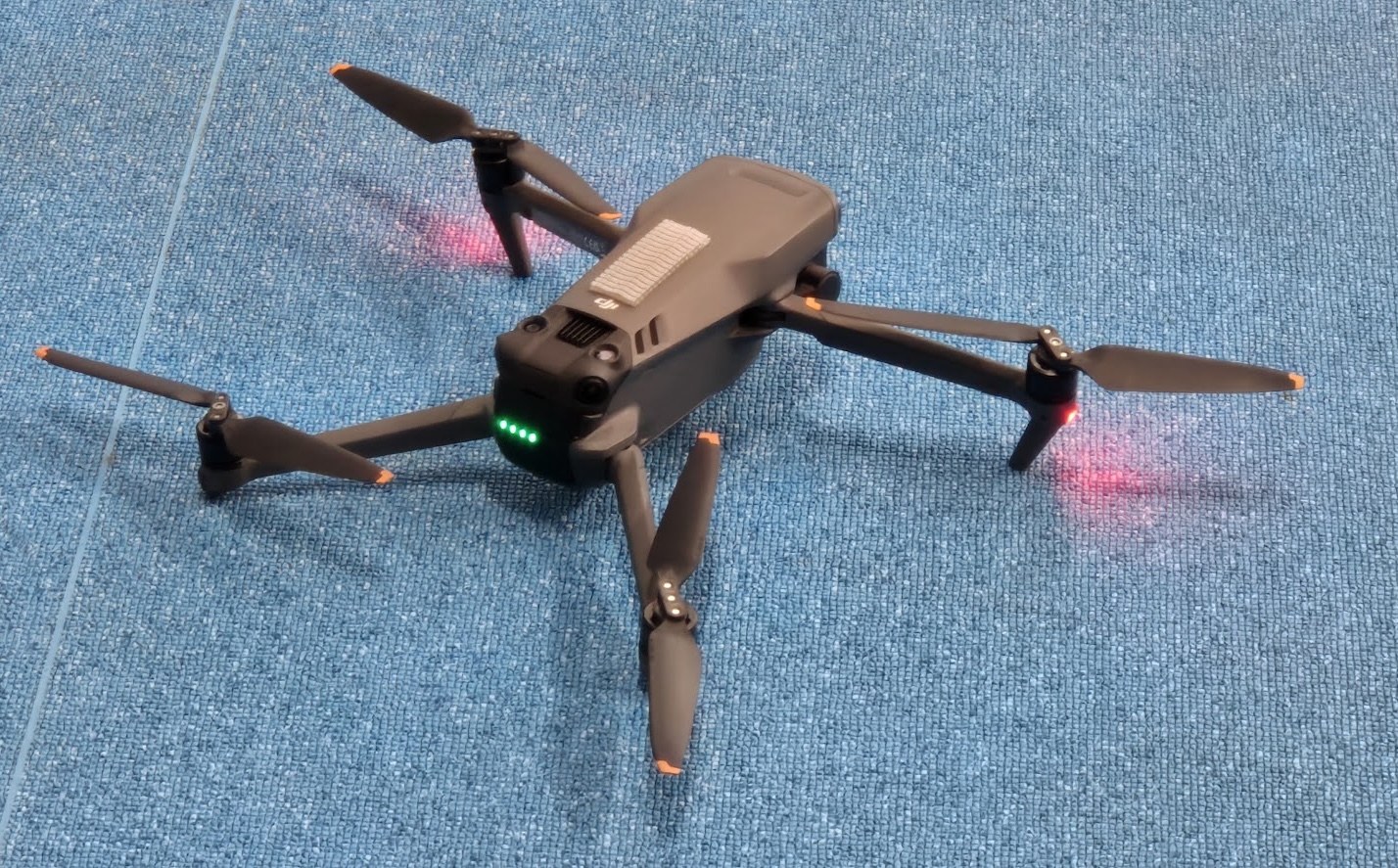}  
  \caption{DJI Mavic 3 Pro}
  \label{fig:drone}
\end{subfigure}
\hspace{0.02\textwidth}
\begin{subfigure}[b]{0.4\textwidth}
  \centering
  \includegraphics[width=1.0\linewidth]{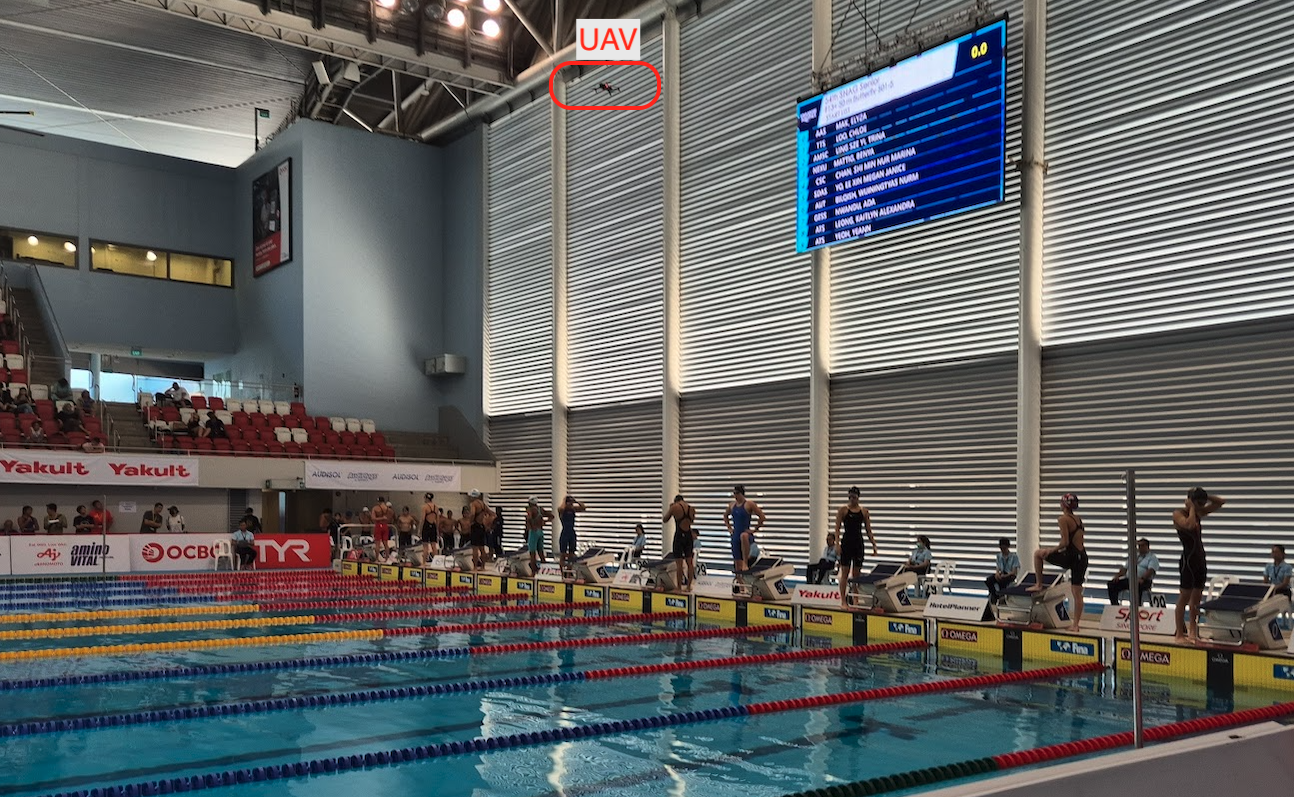} 
  \caption{Data collection during a competition}
  \label{fig:}
\end{subfigure}
\caption{UVA and its demonstration for aerial footage collection}
\label{fig:in_action}
\end{figure*}

We employ a DJI Mavic 3 Pro \cite{dji_mavic3pro}, equipped with a camera that has 4K resolution and operates at 60 fps. The drone is consistently moving at a height of 8~m to get a comprehensive view and ensure that all videos maintain the same scale and perspective throughout the data collection process. Figure \ref{fig:in_action} illustrates this process in a real scenario.

\subsection{Landmark Detection}
The recorded videos then are passed into a body landmark detection model. In this study, we utilized MediaPipe Pose (MPP), an open-source framework developed by Google, to perform human landmark detection on images and videos. MPP employs BlazePose \cite{bazarevsky2020blazepose}, a neural network architecture that can extract and localize 33 key landmarks on the human body. These landmarks are illustrated in \cite{mediapipepose}. Notably, BlazePose has a lightweight design and can deliver real-time inference on mobile devices.


\subsection{Performance Analysis}
In our analysis, we focus on a commonly performed swimming stroke, the \textit{front crawl}. Our system aims to track several essential metrics, including the angles of the upper arms compared to the body centerline, stroke duration and swimming velocity. To achieve this, we leverage specific key points from the 33 landmarks provided by the MPP framework. Specifically, we use landmarks 12-14 and 11-13 to represent the shoulders and landmark 0 to represent the head.

\section{Approach}
\label{sec:approach}
\subsection{Pre-processing}
Swapping between right and left sides is a common issue encountered in current body landmark detection models, as discussed in \cite{giulietti2023swimmernet, zecha2018kinematic}. However, since we focus on front crawl, wherein swimmers face down to execute the strokes properly, we can leverage this information to address left and right side misdetection. For example, if the swimmer moves from left to right, the limbs with larger y-coordinates will correspond to the right side, and if the swimmer moves from right to left, the limbs with larger y-coordinates will correspond to the left side. We use the same idea to determine left and right sides when the swimmer moves up and down.



\subsection{Angles of the upper arms}

\begin{figure*}
  \begin{minipage}{.3\textwidth}
  \centering
    \includegraphics[width=0.7\linewidth]{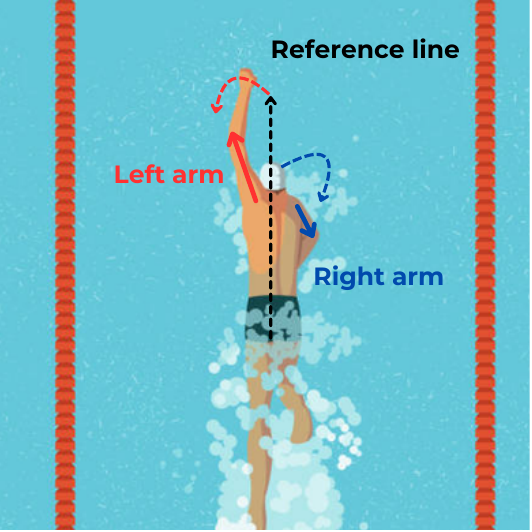}
    \caption{Angles between two arms and the reference line}
    \label{fig:360_angle}
  \end{minipage} \quad
  \begin{minipage}{.3\textwidth}
  \centering
    \includegraphics[width=1.0\linewidth]{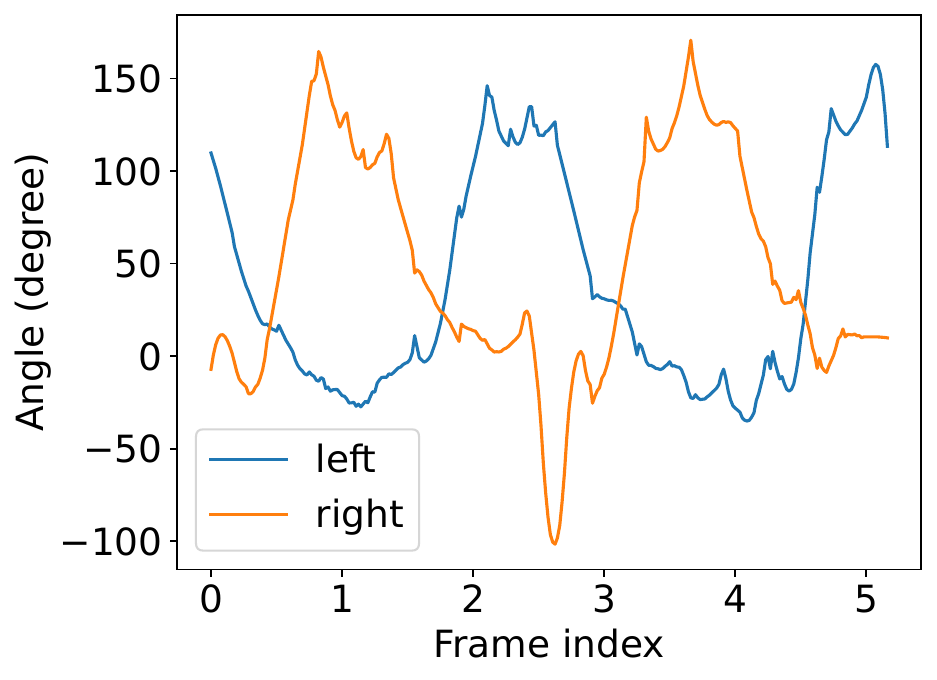}
    \caption{Left and right arm angle in front crawl}
    \label{fig:arm_angle}
  \end{minipage} \quad
  \begin{minipage}{.3\textwidth}
  \centering
    \includegraphics[width=1.0\linewidth]{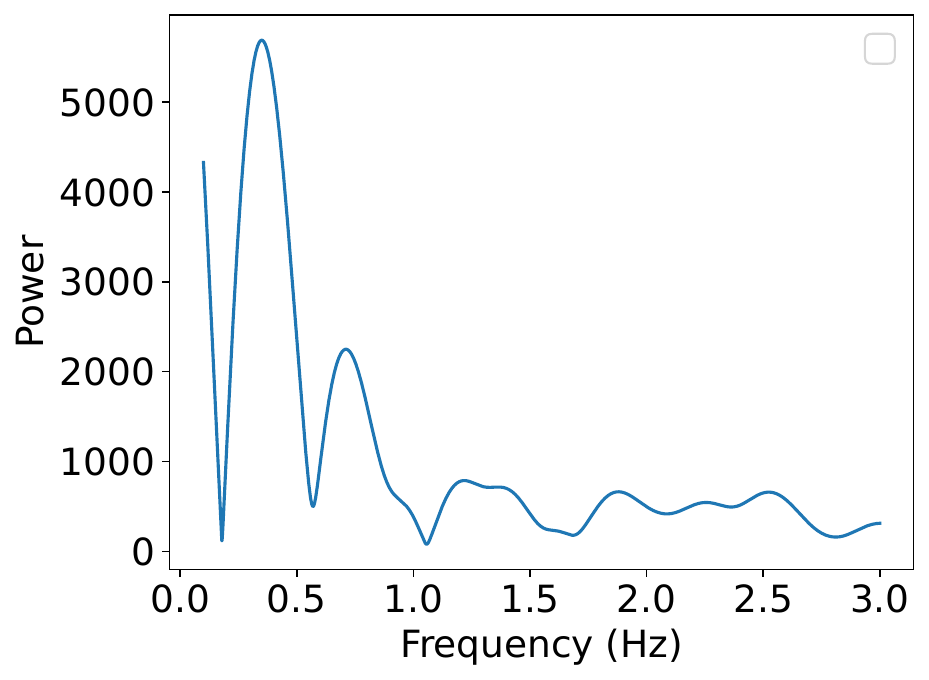}
    \caption{FFT of the right sequence in Figure \ref{fig:arm_angle}}
    \label{fig:fft}
  \end{minipage} \quad
\end{figure*}

Tracking the angles of the upper arm is essential in evaluating swimming performance as it provides insights into the swimmer's technique. For example, by examining the angles on the left and right sides, we can determine if the swimmer's movements are symmetrical. Additionally, this sequence of angles over time allows us to extract the stroke duration. 

To measure this, we first draw a reference line connecting the midpoint of the hips (represented by points 23 and 24) to the nose (represented by point 0). Specifically, we focus on the 360-degree angle formed between the upper arms and the reference line, with the rotation from the swimmer's body, as shown in Figure \ref{fig:360_angle}.

\subsubsection{Symmetry}
Symmetry in swimming reflects overall body balance and technique efficiency \cite{havriluk2007analyzing, carvalho2019water}. 

For years, the symmetry index (SI) has been a quantitative measure used in sports to assess the symmetry or asymmetry of movements. It provides a numerical value that indicates the degree of similarity or dissimilarity between corresponding parameters on the left and right sides of the body. Our study leverages the SI proposed by \cite{robinson1987use}, which is defined as:

\begin{equation}\label{eq:si}
  SI = \frac{X_{right}-X_{left}}{0.5(X_{right}+X_{left})} * 100 
\end{equation}

where $X_{right}$ and $X_{left}$ is the average value of the angle of right and left arms, respectively.

A well-established threshold in literature is that 10\% represents symmetry while higher values indicate asymmetry \cite{jaszczak2008dynamical, sanders2015approach, santos2020symmetry}. The SI can be over 100\% as it is not bounded.


\subsubsection{Stroke duration}
Stroke duration is one of the direct impacts on race time. 
Since all strokes are repetitive, we can extract stroke duration (seconds per cycle) by first applying a Blackman taper function and then a Fast Fourier Transform (FFT) on the sequence of left and right sides. We can deduce the duration of the stroke from the frequency of the signal. In Figure \ref{fig:fft}, the frequency of the right sequence in Figure \ref{fig:arm_angle} is around $0.4 Hz$, which means this swimmer takes  approximately $\frac{1}{0.2}=2.5 seconds$ to complete one cycle of front crawl.

\subsection{Swimming velocity}

\begin{figure}[h]
  \centering
  \includegraphics[width=0.7\linewidth]{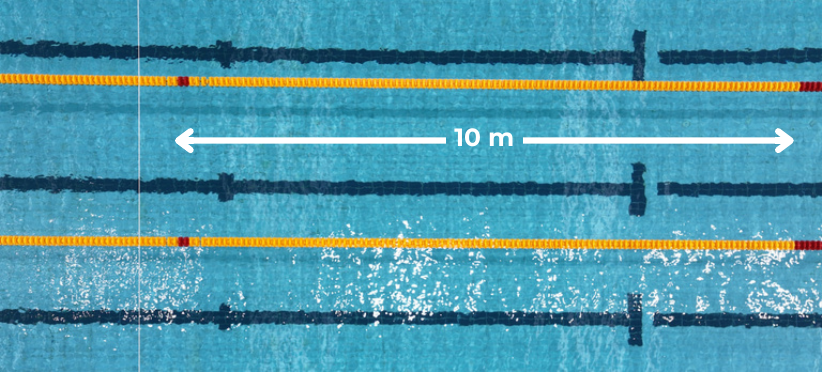}
  \caption{Distances between two markers}
  \label{fig:longcourse}
\end{figure}

To estimate the swimmer's velocity, we make use of the fact that the pool is a standard Olympic-sized pool, where the distance between two consecutive marks is known to be 10 meters. Figure \ref{fig:longcourse} illustrates such a type of measurement for reference. For our study, we record the time at which the swimmer's head reaches each red marker, then calculate her speed based on these time intervals between markers.

\section{Evaluation}
\label{sec:evaluation}
We evaluate our system from five swimmers (3 females (SF), 2 males (SM)) performing front crawl: three of them (SF1, SM1 and SF2) in training  sessions and two of them (SF3 and SM2) in a competition. All perform in a standard Olympic-sized swimming pool. These videos last around 10 to 25 seconds, and the distance they swim ranges from 10 to over 30 meters. 

In our evaluation, we compared swimmers' stroke duration and velocity with ground truth values obtained through visual analysis of the recorded videos. To determine the average stroke duration, we divided the duration of the video by the number of cycles counted within the video. For velocity estimation, we divided the distance (e.g., 10~m, 20~m, or 30~m) by the time it took for the swimmer's head to reach two markers.

\subsection{Detection rate}
The detection rate is calculated as the ratio of the number of detected frames to the total number of frames in the video sequence. The detection rate indicates the model's ability to identify swimming movements under noise and obstructions, such as water reflection and splashes. Table \ref{tab:detection_rate} shows the detection rate of the pose estimation algorithm on our data. During competition, the detection rate drops as more splashes occur.

\begin{table}
  \caption{Detection rate for front crawl}
  \label{tab:detection_rate}
  \begin{tabular}{l c c}
    \toprule
    Mode & Swimmer & Detection rate\\
    \midrule
    Training & \makecell{SF1 \\ SM1 \\ SF2} & \makecell{0.9818 \\ 1.0000 \\ 0.9483 }\\
    \midrule
    Competing & \makecell{SF3 \\ SM2} & \makecell{0.8799 \\ 0.5616}\\
  \bottomrule
\end{tabular}
\end{table}

\subsection{SI and stroke duration}

\begin{figure*}[ht]
\begin{subfigure}[b]{1.\textwidth}
  \centering
  \includegraphics[width=1.0\linewidth]{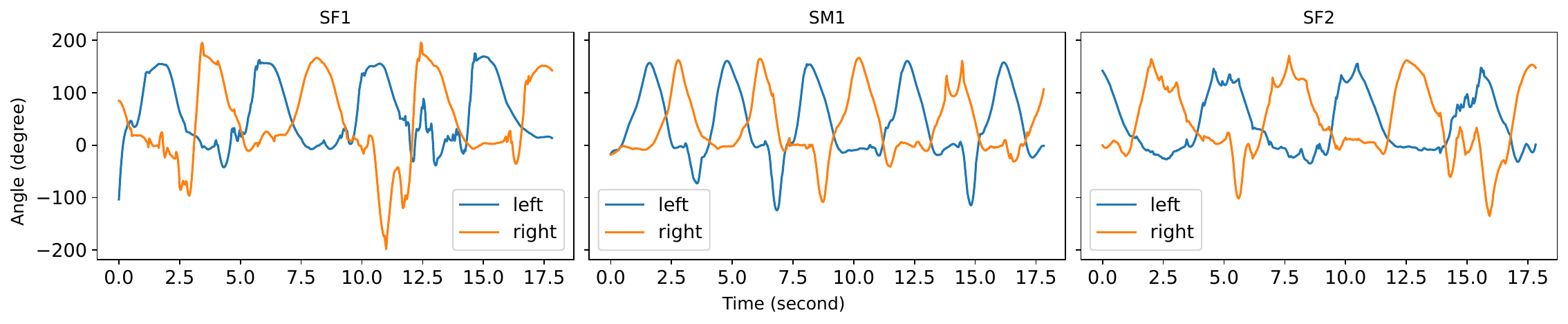}  
  \caption{Arm angle during training}
  \label{fig:training_angle}
\end{subfigure}
\hspace{0.02\textwidth}
\begin{subfigure}[b]{1.\textwidth}
  \centering
  \includegraphics[width=1.0\linewidth]{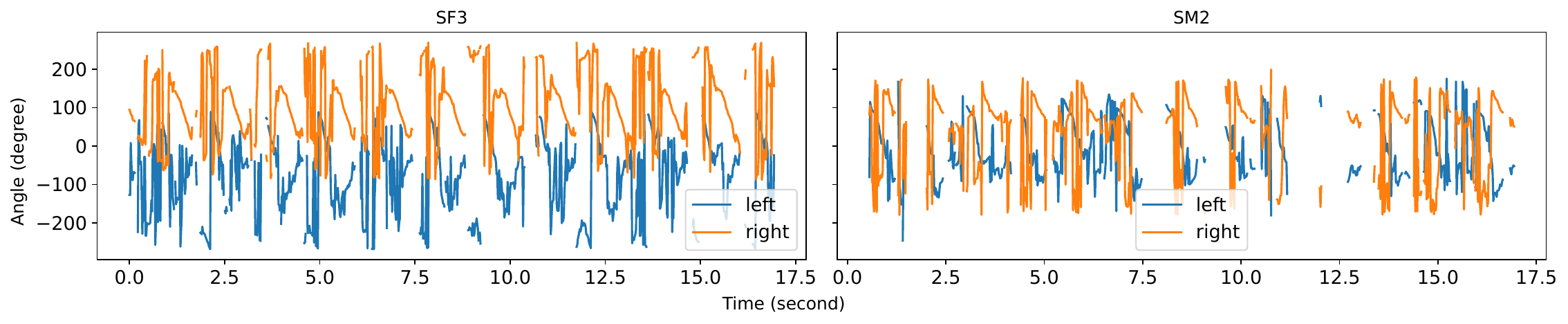} 
  \caption{Arm angle during competition}
  \label{fig:competing_angle}
\end{subfigure}
\caption{Interference removal}
\label{fig:arm_angle_eval}
\end{figure*}

Figure \ref{fig:arm_angle_eval} illustrates the left and right arm angles of the five swimmers. The figure displays unique patterns specific to each swimmer. It also suggests that there are more cycles (i.e. swimming strokes) observed during competition compared to training in the same time unit. However, the number of detected frames is relatively lower during competition. This observation is reasonable as athletes tend to swim at higher speeds and create more splashes during competitive events. These factors pose challenges for the pose estimation model, leading to fewer frames being successfully detected.

\subsubsection{SI}
The SI for each swimmer was calculated using Equation~\ref{eq:si}, and the results are presented in Table~\ref{tab:all_metrics}. We observe that SM1 and SF2 exhibit the most symmetric movements, as their SI values are closest to the cutoff of 10\%. SF1, despite being in the training phase and having a high detection rate, still displays a relatively high SI. This can be attributed to some errors in detection, which introduce inaccuracies in the SI calculation. SF3 and SM2, recorded during competition with increased disturbance, also demonstrate high SI values. These high SI values may be attributed to missed detection during the race.


\begin{table*}
  \caption{Metrics for swimming analysis}
  \label{tab:all_metrics}
  \begin{tabular}{c c c c c c}
    \toprule
    \multirow{2}{4em}{Swimmer} & \multirow{2}{4em}{SI (\%)} & \multicolumn{2}{c}{Stroke duration (s)} & \multicolumn{2}{c}{Swimming velocity (m/s)} \\
    & & Detected & Ground truth & Detected & Ground truth \\
    \midrule
    SF1 & 30.9220 & 1.7840 & 2.0625 & 0.9772 & 1.3333 \\ 
    SM1 & 11.3418 & 1.6254 & 1.7500 & 1.5014 & 1.5385 \\
    SF2 & 11.8773 & 2.6123 & 2.3333 & 1.4106 & 1.2903 \\ 
    \midrule
    SF3 & 26.3405 & 1.5408 & 1.3810 & 1.4883 & 1.4286\\
    SM2 & 101.7017 & 1.0593 & 1.2500 & 1.8437 & 2.0000 \\
  \bottomrule
\end{tabular}
\end{table*}

\subsubsection{Stroke duration}
Table \ref{tab:all_metrics} presents the stroke duration of the five swimmers. Since detection rate is lower during competition, we note that for SF3 and SM2, it is not possible to get satisfactory results when applying FFT. Therefore, we employ peak detection to count the total number of peaks, with the constraint that two peaks must not be within 0.5 seconds of each other to mitigate noise. We then estimate the average stroke duration by dividing the video duration by the number of peaks. The table also demonstrates that the proposed method exhibits an error of less than 0.3 seconds compared to the ground truth for all cases.

\subsection{Swimming velocity}
\begin{figure}[h]
  \centering
  \includegraphics[width=0.6\linewidth]{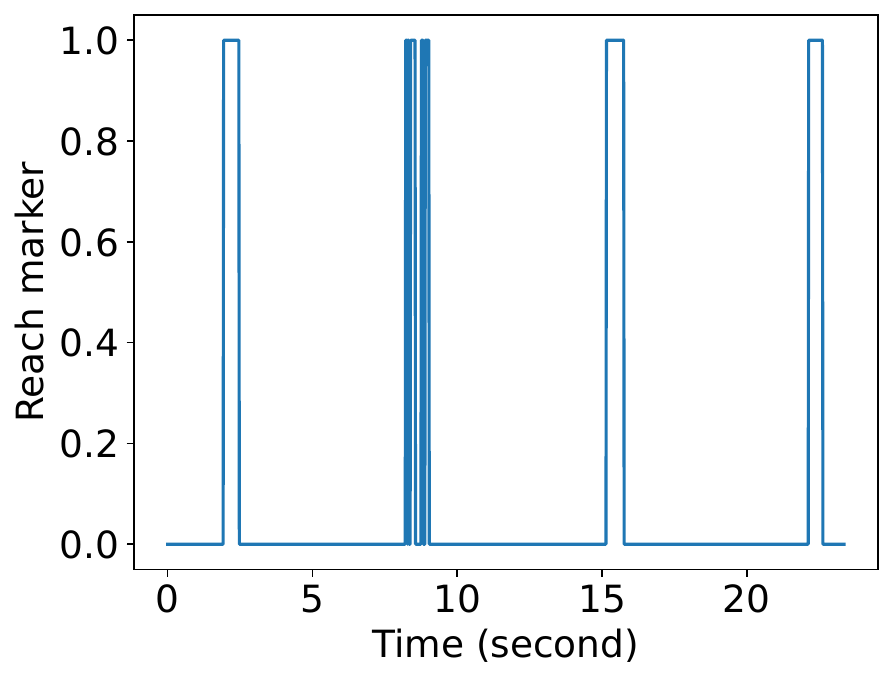}
  \caption{Time and distance marker for SF3}
  \label{fig:marker}
\end{figure}

When the position of the swimmer's head is detected, we check if the left side of that position is also the buoyant marker. We do this by cropping the left side of the image from the head and filtering the marker color. The color of the marker may vary and needs to be defined before running the algorithm. Figure \ref{fig:marker} illustrates the swimming time for SF3, where the y-axis value of 1 indicates the time at which the swimmer's head reaches the marker. According to this figure, SF3 has travelled more than 30 meters. By applying the same methodology to the other swimmers, the swimming velocities for each swimmer can be obtained, as shown in Table \ref{tab:all_metrics}. The table also shows that the errors range from 0.04 to 0.35~m/s. 

\section{Discussion}
\label{sec:discussion}

While UAVs offer a comprehensive view of the pool, they may still encounter occlusions from splashes. These occlusions have a direct impact on the detection rate and the accuracy of pose estimation, affecting other metrics such as stroke duration and swimmer velocity.

Apart from occlusions, the underwater environment brings additional difficulties to the pose estimation model. Body parts submerged in water experience significant distortion caused by water movement and color variations. Further, the dislocation between body parts above and below the water surface due to refraction when light crosses the two mediums makes it difficult for the model to detect an overall human figure.

In our future work, we plan to address these challenges by implementing methods to mitigate splashes, reconstructing accurate body shapes and colors underwater, and fine-tuning the pose estimation model to achieve enhanced accuracy.
\section{Conclusion}
\label{sec:conclusion}
The proposed system offers a novel and effective approach for monitoring swimmer performance. By leveraging a UAV and computer vision algorithms, the system provides comprehensive coverage and unobstructed view. Experiments demonstrated the system's performance in tracking swimmers’ movements, limb angles, stroke duration and velocity in both training and competing settings, with the maximum error of 0.3 seconds for stroke duration and 0.35~m/s for velocity.

This technology has the potential to revolutionize swimmer performance analysis, enabling coaches and athletes to identify areas for improvement, optimize training programs, and enhance overall athletic performance. Future work will focus on improving the system's robustness and exploring advanced performance analysis applications.

\section*{Acknowledgements}
This research was supported by the Singapore Ministry of Education (MOE) Academic Research Fund (AcRF) Tier 1 grant.
\bibliographystyle{ACM-Reference-Format}

\bibliography{reference}
\end{document}